\RequirePackage{xcolor} % try to fix warnings
\documentclass[journal,twoside,web]{ieeecolor}

\usepackage{etoolbox}
\makeatletter
\@ifundefined{color@begingroup}%
  {\let\color@begingroup\relax
   \let\color@endgroup\relax}{}%
\def\fix@ieeecolor@hbox#1{%
  \hbox{\color@begingroup#1\color@endgroup}}
\patchcmd\@makecaption{\hbox}{\fix@ieeecolor@hbox}{}{\FAILED}
\patchcmd\@makecaption{\hbox}{\fix@ieeecolor@hbox}{}{\FAILED}

\usepackage{tmi}
\usepackage{cite}
\usepackage{amsmath,amssymb,amsfonts}
\usepackage{algorithmic}
\usepackage{graphicx}
\usepackage{textcomp}

\usepackage{orcidlink}
\usepackage{hyperref}

\usepackage{booktabs}

\def\BibTeX{{\rm B\kern-.05em{\sc i\kern-.025em b}\kern-.08em
    T\kern-.1667em\lower.7ex\hbox{E}\kern-.125emX}}
\begin{document}

\title{Cross-Modality Image Registration using a Training-Time Privileged Third Modality}

\author{Qianye Yang\,\orcidlink{0000-0003-4401-5311},%\IEEEmembership{Member, IEEE},
        David Atkinson\, \orcidlink{0000-0003-1124-6666},%\IEEEmembership{Member, IEEE},
        Yunguan Fu\,\orcidlink{0000-0002-1184-7421},%\IEEEmembership{Member, IEEE}, 
        Tom Syer\,\orcidlink{0000-0001-9594-9638},%\IEEEmembership{Member, IEEE}, 
        Wen Yan\,\orcidlink{000-0002-3962-5994},%\IEEEmembership{Member, IEEE}, 
        Shonit Punwani\,\orcidlink{0000-0002-1014-0870},%\IEEEmembership{Member, IEEE}, 
        Matthew J. Clarkson\,\orcidlink{0000-0002-5565-1252},%\IEEEmembership{Member, IEEE}, 
        Dean C. Barratt\,\orcidlink{0000-0003-2916-655X},%\IEEEmembership{Member, IEEE}, 
        Tom Vercauteren\,\orcidlink{0000-0003-1794-0456},%\IEEEmembership{Member, IEEE}, 
        Yipeng Hu\,\orcidlink{0000-0003-4902-0486}%\IEEEmembership{Member, IEEE}, 
%\thanks{Manuscript received XX,XX,XXXX; revised XX,XX,XXXX; accepted XX,XX,XXXX. Date of publication XX,XX,XXXX; date of current version XX,XX,XXXX.}
%
\thanks{Q. Yang, M. J. Clarkson, D. C. Barratt and Y. Hu are with the UCL Centre for Medical Image Computing, Department of Medical Physics and Biomedical Engineering, University College London, London WC1E 6BT, U.K., and also with the Wellcome/EPSRC Centre for Interventional and Surgical Sciences University College London, London WC1E 6BT, U.K. (e-mail: qianye.yang.19@ucl.ac.uk; m.clarkson@ucl.ac.uk; d.barratt@ucl.ac.uk; yipeng.hu@ucl.ac.uk).}
\thanks{D. Atkinson is with the Centre for Medical Imaging, University College London, London W1W 7TS, U.K. (e-mail:d.atkinson@ucl.ac.uk)}
\thanks{Y. Fu is with the UCL Centre for Medical Image Computing, Department of Medical Physics and Biomedical Engineering, University College London, London WC1E 6BT, U.K., and also with the InstaDeep Co. (e-mail:yunguan.fu.18@ucl.ac.uk)}
\thanks{T. Syer is with the Centre for Medical Imaging, Division of Medicine, University College London, London WC1E 6BT, U.K. (e-mail:t.syer@ucl.ac.uk)}
\thanks{W. Yan is with City University of Hong Kong, department of Electrical Engineering, Hong Kong, China, and  with the UCL Centre for Medical Image Computing, Department of Medical Physics and Biomedical Engineering, University College London, London WC1E 6BT, U.K. (e-mail:wenyan6-c@my.cityu.edu.hk)}
\thanks{S. Punwani is with the Centre for Medical Imaging, Division of Medicine, University College London, London WC1E 6BT, U.K. (e-mail:s.punwani@ucl.ac.uk)}
\thanks{T. Vercauteren is with the School of Biomedical Engineering \& Imaging Sciences, King's College London, London, U.K. (e-mail:tom.vercauteren@kcl.ac.uk)}%
}

\maketitle

\begin{abstract}
In this work, we consider the task of pairwise cross-modality image registration, which may benefit from exploiting additional images available only at training time from an additional modality that is different to those being registered. As an example, we focus on aligning intra-subject multiparametric Magnetic Resonance (mpMR) images, between T2-weighted (T2w) scans and diffusion-weighted scans with high b-value (DWI$_{high-b}$). For the application of localising tumours in mpMR images, diffusion scans with zero b-value (DWI$_{b=0}$) are considered easier to register to T2w due to the availability of corresponding features. We propose a learning from privileged modality algorithm, using a training-only imaging modality DWI$_{b=0}$, to support the challenging multi-modality registration problems. We present experimental results based on 369 sets of 3D multiparametric MRI images from 356 prostate cancer patients and report, with statistical significance, a lowered median target registration error of 4.34 mm, when registering the holdout DWI$_{high-b}$ and T2w image pairs, compared with that of 7.96 mm before registration. Results also show that the proposed learning-based registration networks enabled efficient registration with comparable or better accuracy, compared with a classical iterative algorithm and other tested learning-based methods with/without the additional modality. These compared algorithms also failed to produce any significantly improved alignment between DWI$_{high-b}$ and T2w in this challenging application.
\end{abstract}


\section{Introduction}
\label{sec:introduction}
\IEEEPARstart{M}{ultiparametric} Magnetic Resonance (mpMR) imaging is now recommended by international guidelines for the initial detection of prostate cancer for men with suspected disease \cite{mottet2021eau,bjurlin2020update,fulgham2017aua}. Most subtypes of prostate cancer diagnosed on mpMR manifest themselves as low signal on T2-weighted MRIs, apparent diffusion coefficient (ADC) images, and high signals on high b-value diffusion MRI (DWI). As shown in both recent radiological and technical studies \cite{delongchamps2011multiparametric,kim2007value,vargas2011diffusion}, mpMR images can lead to more accurate results on prostate cancer detection and staging, compared to only using single modality MR imaging \cite{yang2017co,yang2017joint,aldoj2020semi,cao2019joint,le2017automated}, in particular, the T2-weighted and the diffusion-weighted scans have been recommended as two necessary modalities to include in any mpMR examination\cite{piradsv21}. Jointly assessing mpMR scans can usually be expedited by accurate alignment \cite{yang2017co,giannini2015fully}, especially when localisation of the pathological regions has become increasingly important for followup monitoring, diagnosis and treatment. In real-world clinical data, spatial differences exist between mpMR scans which are usually caused by patient movement during image acquisition, internal organ movement and distortions due to imperfect magnetic fields during image acquisition. However, registering multimodal mpMR images that are designed to provide complementary information is challenging. As these factors are difficult to decouple between different scans, scanner coordinates are often the only geometric reference after acquisition-time magnetic-field correction \cite{rakow2015prostate,embleton2010distortion}. 

For instance, echo-planar imaging using a high diffusion weighting (DWI$_{high-b}$) is considered sensitive for detecting prostate lesions in both peripheral and central gland, but it also can spatially differ from the T2-weighted (T2w) scans, the latter of which provides not only spatial reference for localising tumour of interest but also significant diagnostic value\cite{rosenkrantz2013prostate}.
In many cases, it is visibly evident that registration between the two is required due to the coupled distortion and unknown patient/organ motion. The low signal-to-noise (SNR) in DWI$_{high-b}$ and lack of spatially-corresponding features between the two challenges this registration task for both classical algorithms and recent deep-learning-based methods. Feature-based or semi-automated registration methods have been proposed for this task \cite{de2011fully,fu2021deformable}. In Section~\ref{sec:results}, we provide quantitative evidence to demonstrate the need and the difficulty in direct registration between DWI$_{high-b}$ and T2w scans.

\begin{figure}[tp]
\includegraphics[width=1.0\columnwidth]{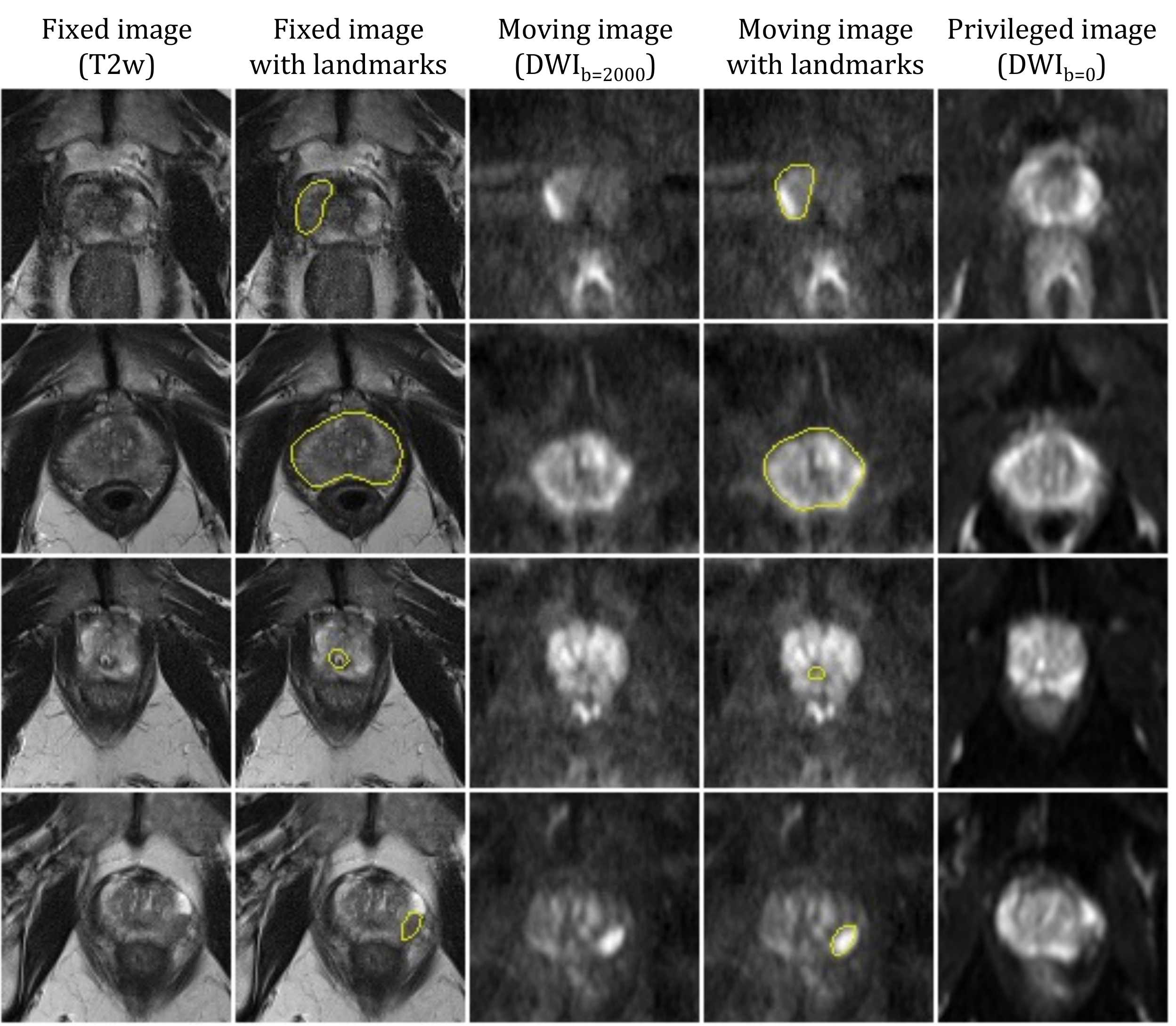}
\caption{Four example cases of the T2w, DWI$_{b=2000}$, and DWI$_{b=0}$ images used in this study. It shows that, compared with DWI$_{b=2000}$, the DWI$_{b=0}$ images in general have richer contrast between different structures and higher signal-to-noise ratio. The misalignment between DWI$_{b=2000}$ and DWI$_{b=0}$ is also smaller than that between DWI$_{b=2000}$ and T2w images. The yellow contours indicate the annotated anatomical landmarks for validation purposes, including tumors, urethra, prostate glands and its zonal structures.}
\label{fig:landmarks}
\end{figure}

DWI scans with low b-value (DWI$_{low-b}$), on the other hand, are less frequently used directly for diagnosis due to its diminished added clinical benefit with the presence of both DWI$_{high-b}$ and T2w scans. For example, a time-critical imaging protocol for high-throughput application, e.g. \cite{full}, may suggest excluding ADC maps, which normally requires DWI$_{low-b}$ to calculate. However, DWI$_{low-b}$ scans are in general of higher SNR than DWI$_{high-b}$ and has better tissue contrast, closer to T2w scans, as shown in Fig.~\ref{fig:landmarks}. Meanwhile, DWI$_{low-b}$ and DWI$_{high-b}$ scans share similar distortion patterns and we have also observed a smaller spatial difference between DWI scans with different b-values, compared to the difference between DWI and T2w scans \cite{de2016diffusion}. Quantitative results for supporting this observation are reported in Section ~\ref{sec:results}. This is probably because DWI$_{low-b}$ is less prone to artifacts and distortion. In this study we use DWI$_{b=0}$ as an example of DWI$_{low-b}$ to facilitate the registration between T2w and DWI$_{high-b}$ images. Although for some simplified imaging protocols, DWI$_{b=0}$ may be omitted, they can still be acquired readily for study purposes, such as neural network training. In addition, DWIs with b-values within a range of 0-100 $sec/mm^2$ can be used as alternatives for DWI$_{b=0}$ images, as suggested by \cite{piradsv21}. In fact, in many existing mpMR imaging protocols for prostate cancer, DWI$_{b=0}$ data have been available for model training purposes. In this study, the DWI$_{b=0}$ data are used as the privileged information for training registration models, which are not required at the inference stage. The possibility and the potential of using other diffusion scans with low b-values may also be interesting under different clinical context, but will not be discussed further in this work.

This work has thus been motivated by a) the above-described clinical scenarios that can take advantage of DWI$_{high-b}$ and T2w, bi-parametric imaging; and b) the hypothesized benefits of using DWI$_{low-b}$ in aiding the cross-modality registration. We investigate deep learning algorithms that incorporate the DWI$_{b=0}$ scans in training registration networks that, once trained, take only DWI$_{high-b}$ and T2w images as network input to register them - a case of learning using privileged information \cite{vapnik2015learning,8080246}. In Section~\ref{sec:methods}, we describe a training strategy to facilitate the use of such a privileged third modality; then compare its performance to the alternative learning-based and non-learning methods; we report in Sect.~\ref{sec:results} experimental results using independent landmarks identified on holdout image pairs of registration interest in this work, i.e. DWI$_{high-b}$ and T2w scans.

The aim of this work is to develop new learning methodologies and test their feasibility in improving the registration performance by incorporating extra imaging modality only in training. Learning-based registration methods have been proposed \cite{hu2018weakly,balakrishnan2019voxelmorph,hering2019enhancing,sun2018towards,sun2018robust,sun2018deformable,stergios2018linear,sokooti2017nonrigid,so2017novel,sloan2018learning,simonovsky2016deep}, especially, taking advantages of highly efficient deep registration networks during inference, with or without graphic processing units (GPUs). Learning-based registration, due to their being formulated as a machine learning task, can readily accommodate other observed latent variables to model additional information, such as a privileged third modality that is of interest in this study.   

The work aims to show quantitative registration results on real clinical data and also highlight that the proposed methods utilising privileged images in this prostate cancer imaging application. However, we also envisage that this type of algorithms may be of wider applicability to other medical image registration problems. For example, longitudinal image registration when training data are available at more time points from retrospective subjects than those that need registration, or an interventional image registration task that with a missing reference image that is easier to register due to larger field-of-view or better image quality. The experiments presented in this work is focusing on unsupervised registration \cite{qin2019unsupervised,balakrishnan2018unsupervised,krebs2018unsupervised,qin2018joint,dalca2018unsupervised}, due to the challenges in identifying substantial number of corresponding regions of interest (ROIs) labels for weak supervision \cite{hu2018weakly,balakrishnan2019voxelmorph,hering2019enhancing}. Other approaches using deep feature for multi-modal registration \cite{lee2019image}, also requires anatomical annotations to learn registration-useful representations. However, when such labels are available, they may further aid cross-modality registration with the additional modality, but may be considered outside of the scope of this work.

We summarise the contributions in this work: 1) we propose to use additional images only in training to assist a challenging mpMR image registration task; 2) we propose and compare registration network training strategies using the privileged images; 3) we present experimental results using clinical imaging data from 356 prostate cancer patients; and 4) we provide quantitative results comparing the proposed methods with other learning-based registration methods with and without using the privileged third modality, in addition to a comparison to a non-learning algorithm, and report improved or non-inferior registration performance from the proposed registration algorithm.

\section{Methods}
\label{sec:methods}
In this section, we describe a training strategy to train a registration network $f^{{M}\rightarrow{F}}_\theta(\textbf{X}^{M},\textbf{X}^{F})$ with network parameters $\theta$ and to input moving and fixed image pair $(\textbf{X}^{M},\textbf{X}^{F})$, given a set of training image trios $\{(x^{M}_{n}, x^{F}_{n}, x^{P}_{n}), n=1,2,...N\}$, where $N$ is the total number of MR studies. $x^{M}_{n}$, $x^{F}_{n}$ and $x^{P}_{n}$ are moving, fixed, and privileged images available during training, respectively. The registration network $f_\theta$ takes only two images as input and predicts the transformation, e.g. $\mu_n^{{M}\leftarrow{F}}=f^{{M}\rightarrow{F}}_\theta(x^{M}_{n},x^{F}_{n})$, where $\mu_n^{{M}\leftarrow{F}}$ is a dense displacement field (DDF) that can be used to obtain the warped moving image $x^{M}_{n} \circ \mu_n^{{M}\leftarrow{F}}$, where $\circ$ represents the resampling operation.

\subsection{Learning from privileged supervision}
\label{sec:methods.adapted}
First, we describe a formulation that enables training registration networks using the third modality that is not required during inference, as the network does not take the third image modality $x^{P}_{n}$ as input, but rather is considered as a special type of supervision.

\begin{figure*}
\centerline{\includegraphics[width=2\columnwidth]{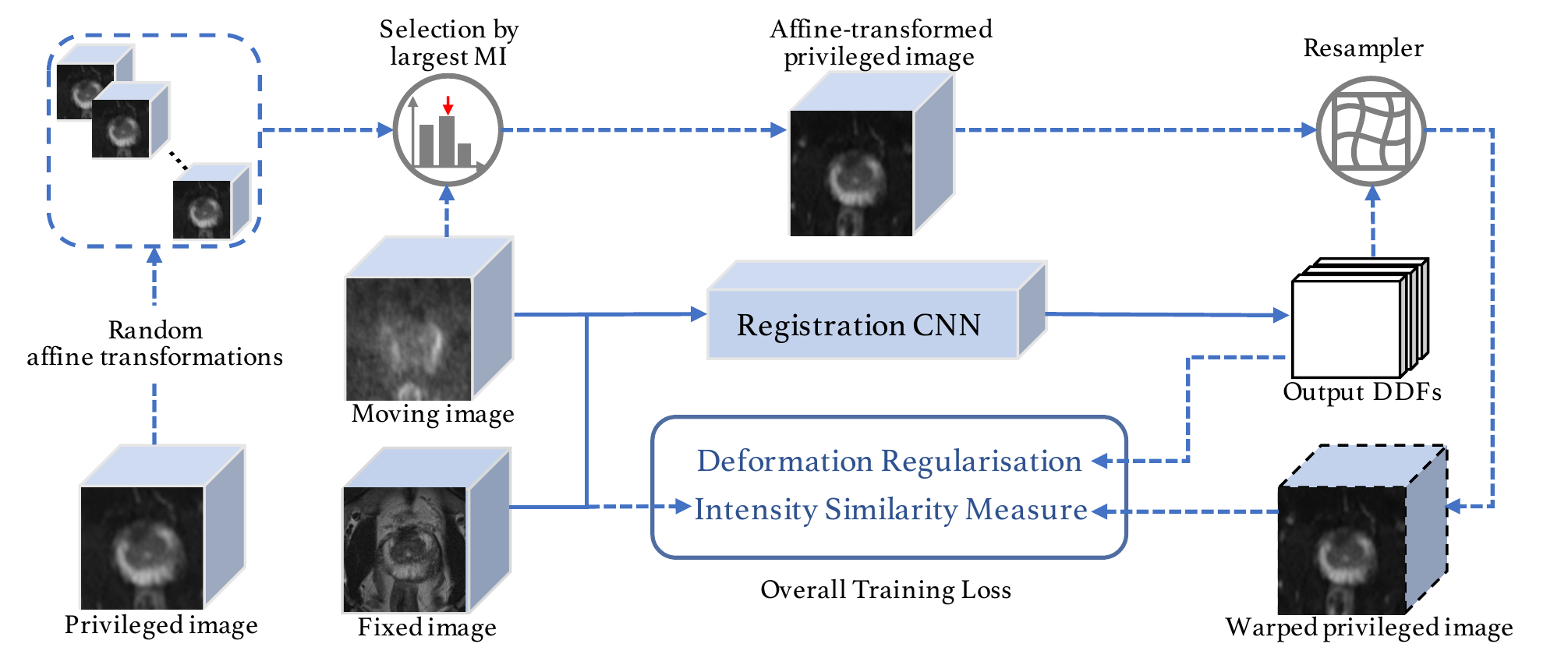}}
\caption{The proposed privileged supervision for training a registration network in Section~\ref{sec:methods.adapted}. The dotted lines indicate the data flow only used in training.}
\label{fig:methods.adapted}
\end{figure*}

This is conceptually similar to weak supervision \cite{hu2018weakly}, where the segmentation labels of regions of interest have been proposed for weakly supervising registration networks. As illustrated in Fig.~\ref{fig:methods.adapted}, the proposed registration network $f^{{M}\rightarrow{F}}_\theta$ accepts the same input image pairs, $x^{M}_{n}$ and $x^{F}_{n}$, but is trained by maximising the image similarity between warped privileged images $x^{P}_{n} \circ \mu_n^{{M}\leftarrow{F}}$ and fixed images $x^{F}_{n}$, as opposed to the similarity measure used in an unsupervised approach between warped moving images $x^{M}_{n} \circ \mu_n^{{M}\leftarrow{F}}$ and the fixed images. 

This formulation can also be considered as using the privileged-image-generated DDFs $\hat{\mu}_n^{{P}\leftarrow{F}}$ as the noisy labels for $\mu_n^{{M}\leftarrow{F}}$. It is important to highlight that the necessary condition for an unbiased estimate of $\mu_n^{{M}\leftarrow{F}}$ is $\mathbb{E}[\mu_n^{{M}\leftarrow{F}}]=\mathbb{E}[\hat{\mu}_n^{{P}\leftarrow{F}}]$, rather than the stringent sufficient condition $\mu_n^{{M}\leftarrow{F}}=\hat{\mu}_n^{{P}\leftarrow{F}}, \forall n$, since the method aims to provide a good estimate of the expected (average) of the deformation among all image pairs, rather than precise estimation for individual training image pairs.

\subsection{Monte-Carlo resampling for bias reduction}
\label{sec:methods.surrogate}
Second, we develop a simple yet effective numerical resampling procedure to maximise the benefits of the privileged images during registration network training described in Section~\ref{sec:methods.adapted}. 

To reduce the bias $\mathbb{E}[\mu_n^{{M}\leftarrow{F}}-\hat{\mu}_n^{{P}\leftarrow{F}}]=\mathbb{E}[\mu_n^{{M}\leftarrow{F}}]-\mathbb{E}[\hat{\mu}_n^{{P}\leftarrow{F}}]$, it is sufficient to spatially align the moving and the privileged images, $x^{M}_{n}$ and $x^{P}_{n}$, which results in algorithms that are similar to the joint training (Section~\ref{sec:methods.joint}). As argued earlier, aligning $x^{M}_{n}$ and $x^{P}_{n}$ is itself a multimodal image registration that can be challenging or unreliable in practice. We propose a simple Monte Carlo update step\cite{liu2001theoretical} to reduce the upper-bound of this bias, using affine-transformed privileged images $\tilde{x}^{P}_{n}$ before being warped by the network-generated DDFs $\mu_n^{{M}\leftarrow{F}}$. Here, $\tilde{x}^{P}_{n}=\operatorname{argmax}_{x\in \{x^{P}_{n} \circ A_i\}}[MI(x^{M}_{n}, x)]$, where $\{A_i\} _{i=1}^I$ is a set of $I=5$ randomly generated affine transformations. A proof of this bias upper-bound is provided in the following section. The ``standard'' unsupervised loss is used here between the warped privileged images and the fixed images, with a weighted deformation regularisation term $\mathcal{C}$.

\begin{equation}
    J(\theta) = -\alpha\cdot MI(x^{F}_{n}, \tilde{x}^{P}_{n} \circ \mu_n^{{M}\leftarrow{F}}) + \beta\cdot\mathcal{C}(\mu_n^{{M}\leftarrow{F}})
\label{eq:loss2.3}
\end{equation}

\subsection{Effectiveness of surrogate supervision}
\label{sec:methods.proof}
Here, we provide an analysis to show that the Monte-Carlo procedure, described in Section~\ref{sec:methods.adapted}, is effective to reduce the bias between the warped privileged image and the ground-truth image without accessing to the ground-truth.

Denote a set of training image trios $(x^{M}, x^{F}, x^{P})\in\mathcal{X}^3$, representing the moving, fixed and privileged images, respectively. $\mathcal{X}$ is the vector space for images.
Given a dense displacement field $\mu^{M\leftarrow{F}}$, denote $T:\mathcal{X}\rightarrow{\mathcal{X}}$ that maps $x^M$ to $T(x^M) = x^M \circ \mu^{M\leftarrow{F}}$ and $\circ$ represents resampling.

The registration task is thereby to minimize the difference between the warped moving image and the fixed image:
\begin{align}
    J = d( x^{M} \circ \mu^{{M}\leftarrow{F}}, x^F),
\end{align}
where $d:\mathcal{X}^2\rightarrow{[0,+\infty)}$ is a metric defined on $\mathcal{X}$.

The proposed method uses an affine transformed privileged image $\tilde{x}^P$ as the surrogate of $x^M$, therefore it minimizes a different objective $J_\text{surrogate}$:
\begin{align}
    J_\text{surrogate} = d( \tilde{x}^{P} \circ \mu^{{M}\leftarrow{F}}, x^F),
\end{align}

Using triangulation inequality \cite{fitzpatrick2007euclid}, the difference between the two objectives has the following upper-bound:
\begin{align*}
    J_\text{surrogate}- J =~& d( \tilde{x}^{P} \circ \mu^{{M}\leftarrow{F}}, x^F) - d( x^{M} \circ \mu^{{M}\leftarrow{F}}, x^F)\\
    \leq~& d( \tilde{x}^{P} \circ \mu^{{M}\leftarrow{F}},  x^{M} \circ \mu^{{M}\leftarrow{F}}) \quad \\
    =~& d(T(\tilde{x}^{P}),~T(x^M))
\end{align*}

If $T$ is Lipschitz continuous, then there exists a constant $K$ such that
\begin{align*}
    J_\text{surrogate}- J \leq~ K d( \tilde{x}^{P},  x^{M})
\end{align*}

Thus, the surrogate objective $J_\text{surrogate}$ approximates the target objective $J$, when $d( \tilde{x}^{P},  x^{M})$ is minimised. This justifies the proposed update step, in which multiple random affine transformations are applied on the privileged image and the closest one to the moving image is then selected. However, in this application, the adopted mutual information based distance $d(x_1,x_2) = - MI(x_1;x_2)$ is not strictly a \textit{metric} on $\mathcal{X}$. Complications of the use of MI may warrant further investigation, but in practice, the above-described Monte-Carlo procedure almost always found a resampled images that lower the MI to the moving image with as few as $5$-$10$ samples.

\subsection{Alternative methods for utilising the third modality}
Last but not least, it is important to test other, arguably simpler, approaches for training registration networks that can utilise the privileged information from the latent third modality. We describe two such alternatives as below, in which the third images are used only in training and are not required during inference. 

\subsubsection{Joint training}
\label{sec:methods.joint}
One approach to utilise the privileged images $x^{P}_{n}$ is to estimate ground-truth DDFs $\hat{\mu}_n^{{M}\leftarrow{F}}$, by composing two intermediate transformation, $\hat{\mu}_n^{{M}\leftarrow{F}}=\hat{\mu}_n^{{M}\leftarrow{P}} \circ \hat{\mu}_n^{{P}\leftarrow{F}}$. While either classical algorithms or learning-based registration networks can be used to estimate $\hat{\mu}_n^{{M}\leftarrow{P}}$ and $\hat{\mu}_n^{{P}\leftarrow{F}}$ independent of training $f^{{M}\rightarrow{F}}_\theta$, we discuss two joint training algorithms. 

The first algorithm trains three registration networks, $f^{{M}\rightarrow{F}}_\theta$, $f^{{M}\rightarrow{P}}_{\phi_1}$ and $f^{{P}\rightarrow{F}}_{\phi_2}$, to simultaneously estimate $\hat{\mu}_n^{{M}\leftarrow{F}}$, $\hat{\mu}_n^{{M}\leftarrow{P}}$ and $\hat{\mu}_n^{{P}\leftarrow{F}}$, respectively. A mean-square difference (MSD) can be used to minimise the difference between the network-predicted $\mu_n^{{M}\leftarrow{F}}$ and estimated ground-truth $\hat{\mu}_n^{{M}\leftarrow{P}} \circ \hat{\mu}_n^{{P}\leftarrow{F}}$. To train the latter two networks, an image dissimilarity loss, such as mutual information (MI), can be used between $(x^{P}_{n},x^{M}_{n} \circ \mu_n^{{M}\leftarrow{P}})$ and between $(x^{F}_{n},x^{P}_{n} \circ \mu_n^{{P}\leftarrow{F}})$. 

With feature-rich moving images, a variant of the joint training can be implemented by maximising the image similarity between $x^{M}_{n} \circ \hat{\mu}_n^{{M}\leftarrow{P}} \circ \hat{\mu}_n^{{P}\leftarrow{F}}$ and $x^{M}_{n} \circ \hat{\mu}_n^{{M}\leftarrow{F}}$, without explicitly minimising the loss on the DDF difference.
As the alignment between the two transformed moving images can be effectively measured by MSD. Results presented in this work are based on the following joint training loss in its general form:
\begin{align}
    J(\theta) = & \ J^{M\rightarrow{F}}+J^{M\rightarrow{P}}+J^{P\rightarrow{F}}+ \\  \nonumber
                & \ MSD(x^{M}_{n}\circ\mu_n^{{M}\leftarrow{F}}, x^{M}_{n}\circ\mu_n^{{M}\leftarrow{P}}\circ\mu_n^{{P}\leftarrow{F}})
\label{eq:loss1.2}
\end{align}
where, 
\begin{equation}
    J^{A\rightarrow{B}}(\theta) = -\alpha\cdot MI(x^{B}_{n}, x^{A}_{n}\circ\mu_n^{{A}\leftarrow{B}}) + \beta\cdot\mathcal{C}(\mu_n^{{A}\leftarrow{B}})
\label{eq:loss1.1}
\end{equation}
where, the image similarity and a deformation regularisation term $\mathcal{C}(\mu_n^{{A}\leftarrow{B}})$ are weighted by $\alpha$ and $\beta$, respectively, with shared values between the terms in Eq.1. $L^2$-norm on DDF gradient is used in this work: $\mathcal{C}(\mu_n^{{A}\leftarrow{B}})=||\nabla\mu_n^{{A}\leftarrow{B}}||_{2}$.

\subsubsection{Mixed sampling}
\label{sec:methods.mixed}
Rather than using the $x^{P}_{n}$ as an intermediate imaging modality as in Section~\ref{sec:methods.joint}, we consider to learn a shared registration network to predict the DDFs from both pairs of images, $(x^{M}_{n},x^{F}_{n})$ and $(x^{P}_{n},x^{F}_{n})$. An unsupervised registration network can be trained by sampling moving and fixed image pairs from the mixed set $\{(x^{M}_{n},x^{F}_{n})\} \cup \{(x^{P}_{n},x^{F}_{n})\}$. The loss function is given by:

\begin{align}
    J(\theta) = & \ -\alpha(MI(x^{F}_{n}, x^{M}_{n}\circ\mu_n^{{M}\leftarrow{F}}) + \\ \nonumber
                & \ MI(x^{F}_{n}, x^{P}_{n}\circ\mu_n^{{P}\leftarrow{F}})) + \beta\cdot\mathcal{C}(\mu_n^{(\theta)})
\label{eq:loss2.2}
\end{align}
where, hyper-parameters $\alpha$ and $\beta$ specify the weights on the intensity dissimilarity and deformation regularisation, respectively.

This unsupervised approach utilises $x^{P}_{n}$ during training, but still uses an image dissimilarity measure between transformed moving images and the fixed images. The lack of reliable and robust similarity measure between the two has not been addressed directly. While methods such as domain adaptation and semi-supervised learning make use of similarity between modalities $x^{P}_{n}$, $x^{M}_{n}$ and $x^{F}_{n}$, such that the registration network predict reasonable DDFs without robust measure between $x^{M}_{n}$ and $x^{F}_{n}$. These remain interesting future research, although it might also be further complicated by the distribution shift between training set $\{(x^{M}_{n},x^{F}_{n})\} \cup \{(x^{P}_{n},x^{F}_{n})\}$ and testing set $\{(x^{M}_{n},x^{F}_{n})\}$. Nevertheless, the described mixed sampling presents a reference performance from a single registration network, with quantitative results reported in Section~\ref{sec:results}.

\subsection{Evaluation}
All the registration networks described in Section~\ref{sec:methods} aim to register the moving and fixed images, without using the privileged images at test time. The anatomical and pathological landmarks are manually identified including patient-specific tumors, urethra, prostate glands and zonal structures, and labelled volumetrically as binary masks. The root-mean-square distance was computed as target registration errors (TREs), between the centers of the mass of the corresponding landmarks independently defined on the fixed and network-warped moving images on holdout data set.

Experiment details are described in Section~\ref{sec:exps}, in which the intra-subject DWI$_{high-b}$, T2w and DWI$_{b=0}$ are used as the moving, fixed and privileged images, respectively. When appropriate, MI is also reported, which may be less relevant to the quality of registration, compared to the TREs on independent landmarks, but provides a quantitative measure how the optimisation during training and generalisation during inference perform. When comparisons are made, p-values are reported from paired two-sided t-tests at a significance level of $\alpha=0.05$.

\section{Experiments}
\label{sec:exps}
\subsection{Data and preprocessing}
369 mpMR image studies were acquired from 356 prostate cancer patients at University College London Hospitals. One or two studies of mpMR images were available for each patient. The mpMRIs were acquired from 1.5T SIEMENS MR scanners, with original voxel resolution of $0.625\times0.625\times1.0$ $mm^3$ and $1.0\times1.0\times5.0$ $mm^3$ for the T2w and DWIs, respectively. All the image volumes were resampled to voxel dimension of $1.0\times1.0\times1.0$ $mm^3$ and got a center-cropped volume of $104\times104\times92$ voxels, with a normalised intensity range of $[0,1]$. In order to validate the registration performance, 35 pairs of mpMRIs from 35 patients with obviously large initial misalignment were selected as the holdout set. The rest of the data set was split into 302 and 32 MRI studies, from 289 and 32 patients, for training and validation sets, respectively. Up to three pairs of landmarks were identified for each study and a total of 50 pairs of landmarks were labelled for the holdout set. The annotation of the landmarks was performed by two biomedical imaging researchers, who have completed a BAUS-accredited MRI course on prostate cancer. The landmarks were labelled by one observer before being checked by the other. To investigate the intra-observer variance, the holdout test set was annotated again, two-months after, and blind to, the first annotation. An intra-observer landmark localization error of 1.08$\pm$0.54mm is achieved.

Two additional data set were used for external validation. Data Set A was acquired from a different hospital, with an approved Institutional Review Board protocol designed at the University College London Hospital (UCLH). The original voxel resolution was $0.625\times0.625\times1.0$ $mm^3$ and $2.0\times2.0\times5.0$ $mm^3$ for the T2w and diffusion-weighted images, respectively. Data Set B was obtained from the Cancer Imaging Archive \cite{fedorov2017multiparametric}, with the original voxel resolutions of $0.27\times0.27\times3.0$ $mm^3$ and $0.7\times0.7\times4.0$ $mm^3$, for the T2w and DWIs, respectively. The mpMRIs were acquired from a 3T GE MR scanner, with endorectal coil. In this public data sets, we only have access to the DWI$_{high-b}$ with b=1400 $sec/mm^2$ in this data set. The same image prepossessing and the landmark annotation were used, as on the UCLH data set. A total of 30 patients with 42 pairs of landmarks and 20 patients with 21 pairs of landmarks are used in the Data Sets A and B, respectively, for assessing the registration performance on external data sets.

The MI was adopted for the similarity measure, suggested by a previous study~\cite{chappelow2011elastic}. The MI was also used as the validation metric for hyperparameter search, specifying the weightings of loss terms $\alpha$ and $\beta$ to 0.5 and $1\times10^3$, respectively. Fine-tuning of these hyper-parameters by, for example, systematic or automated hyperparameter search should benefit and is a subject of future studies.

\subsection{Network training}
An encoder-decoder registration network \cite{hu2018weakly} was used for DDF prediction in all the models in this work. Random affine transformations were added to the input of the network, both for data augmentation and the Monte-Carlo resampling. The method of the random affine transformation is adapted from the open-source code DeepReg \cite{Fu2020}, which is generated by randomly resampling the image corners from a uniform distribution, in order to keep minimal sampling outside the original image. The image warping method is implemented using a standard grid sampling method with trilinear interpolation and zero-padding \cite{Fu2020}. The network training was implemented with PyTorch \cite{paszke2019pytorch} and made open-source https://github.com/QianyeYang/mpmrireg. The Adam optimizer with an initial learning rate of $10^{-5}$ was used. The ``privileged supervision'' networks described in Section~\ref{fig:methods.adapted} were trained on Nvidia Tesla V100 GPUs with a minibatch of 4 sets of image data, each containing a trio of intra-subject DWI$_{b=2000}$, T2w and DWI$_{b=0}$ images. Each network was run for 600,000 iterations, approximately 50 hours. All registration networks were trained using the same training strategy unless otherwise specified.

\subsection{Other learning-based registration}
\label{sec:exps.learning}
The ``joint training'' and the ``mixed sampling'' networks that implemented methods described in Section~\ref{sec:methods.joint} and Section~\ref{sec:methods.mixed}, respectively, were also trained to test these alternative approaches to incorporate the third image modality. Like in evaluating the privileged supervision network, these two networks were trained with the trio of intra-subjective images, but only took T2w and DWI$_{b=2000}$ images as input, during test stage using the holdout set. 

In addition, a learning-based registration methods were compared for directly aligning T2w and DWI scans with b values being 2000, DWI$_{b=2000}$. The registration network was trained using the unsupervised learning algorithm, similar to the one used in Section~\ref{sec:methods.mixed}, but with only T2w and DWI$_{b=2000}$ sampled in training without DWI$_{b=0}$. This is referred to as the ``Direct'' method. For further understanding the role of DWI$_{b=0}$ scans and the potential benefits in adding the bias-reducing Monte-Carlo resampling, described in Section.~\ref{sec:methods.surrogate}, another unsupervised registration network was trained using only T2w and DWI$_{b=0}$ in training without DWI$_{b=2000}$. These two registration networks were both tested on registering the T2w and DWI$_{b=2000}$ images on the holdout set.

Weakly supervised registration \cite{hu2018weakly,balakrishnan2019voxelmorph} methods have also been proved to be effective for the multi-modal registration problems. However, the gland masks of the DWI$_{b=2000}$ in this study are not available and arguably much more difficult to annotate accurately. For example, rectal gas is known for generating magnetic distortion around posterior regions of the prostate glands, which complicates in determining capsule boundaries; and DWI$_{high-b}$ has high sensitivity in certain types of pathology but lacks contrast in gland itself. This study is to investigate how much improvement is feasible for using unsupervised learning methods with unlabelled image data, which are in practice more feasible to obtain.

\subsection{Non-learning registration}
Learning-based registration methods in general provide superior efficiency, compared with the alternative classical registration algorithms based on iterative optimisation, especially for large 3D volumetric medical images \cite{nazib2018comparative}. However, it is useful to report the performance using the classical methods which have been developed for registering multimodal image registration in similar applications \cite{pluim2003mutual,gaens1998non,lu2008mutual,maes1997multimodality,sun2021crossmodalnet}, for last two decades. 

For its fast GPU implementation, the NiftyReg package was used to compare a non-rigid B-spline based free-form deformation algorithm with the above learning-based methods, by directly registering the T2w and DWI$_{b=2000}$. The NiftyReg Package was used as an example of non-learning algorithms, with normalised mutual information and other parameter values followed a previous prostate MR registration study \cite{yang2020longitudinal}. In our experiment, MI was used as the similarity measure for comparison purpose with a bending energy weight of 0.005, among other default configurations. These parameters may not be directly comparable to those used in the learning-based algorithms due to difference between the pairwise optimisation and stochastic-gradient-based learning process, in addition to varying implementation choices. The MI values before and after respective algorithms are reported in Sect~\ref{sec:results}.

The aim for reporting results from the non-learning registration is not intended to compare their registration accuracy, as substantially more comprehensive experiments shall be required to draw a convincing conclusion that may also be dependent on the application and the experimental data used. Rather, this provides a reference of the registration performance with a readily-available, non-learning registration algorithm that does not require the third modality in this specific prostate cancer application.

\section{Results}
\label{sec:results}
\subsection{Registration performance on holdout set}\label{sec:results.holdout}
The TRE and MI results on the holdout set are summarised in Table.~\ref{tab:results}. The proposed privileged supervision increased the median MI from 0.06 to 0.20 and lowered the median TRE from 7.96 mm to 4.34 mm, improved from those before registration, both with statistical significance (p-values$<$0.001). All the other tested methods showed improved TREs in this application with statistical significance as well (NiftyReg:p-value=0.03, others:p-value$<$0.001). The Jacobian determinants of each predicted DDF was also computed and, from all proposed methods in this study, no negative values were found.

Results from two groups of cases are also summarised in Table.~\ref{tab:results.top}, with those that have the largest initial misalignment, measured by landmark distance before registration, and the most improvement by registration, measured by TRE. It is noteworthy that the latter group was selected by the improvement after the registration results were obtained, which were not be available prior to registration, therefore only provides a selective reference for measuring the potential contribution. Together with the cases with largest misalignemnt, these two subgroup results represent the comparison on those cases that need the registration the most. 

\begin{figure*}[tp]
\centerline{\includegraphics[width=1.50\columnwidth]{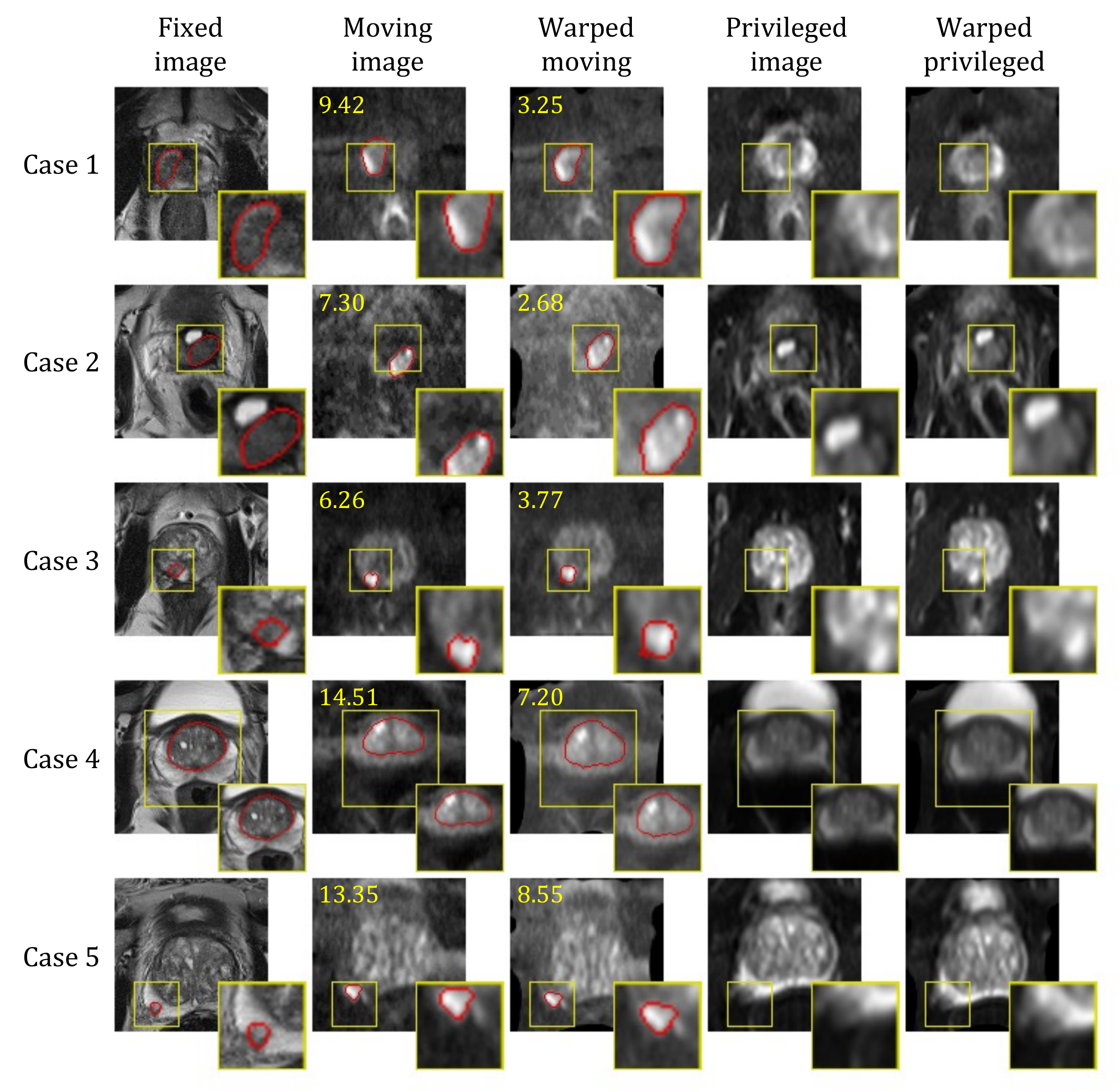}}
\caption{Five example registered cases using the proposed privileged supervision method with yellow bounding boxes, zoomed in at the same reference positions, and the red landmark contours that indicate ROIs for assessing registration. Annotation on privileged images, in the last two columns, are challenging and not available during test time, with images being presented for comparison. For each case, the bounding boxes are placed at the same reference locations in order to assessing the registration. The the TREs (mm) before and after registration are provided in yellow, on the moving and the warped moving images, respectively.}
\label{fig:results}
\end{figure*}

The registration results have been visually assessed and examples are provided in Fig.~\ref{fig:results}. In the first three cases, the morphology and the location of the tumor are more consistent with the fixed T2w images after registration. The fourth and the fifth cases are challenging cases with larger initial misalignment. Case 4 shows a visible improvement in morphology of the central gland. The registration compensated the distorted region near the rectum. Meanwhile, the increased hyper-intensity area on the top of the warped privileged image indicates an improved alignment of the bladder. In case 5, although the registration could be further improved, the registered location of the prostate gland and the tumor indicate the predicted contributed to visibly reduce the misalignment.

\begin{figure*}[tp]
\centerline{\includegraphics[width=2.05\columnwidth]{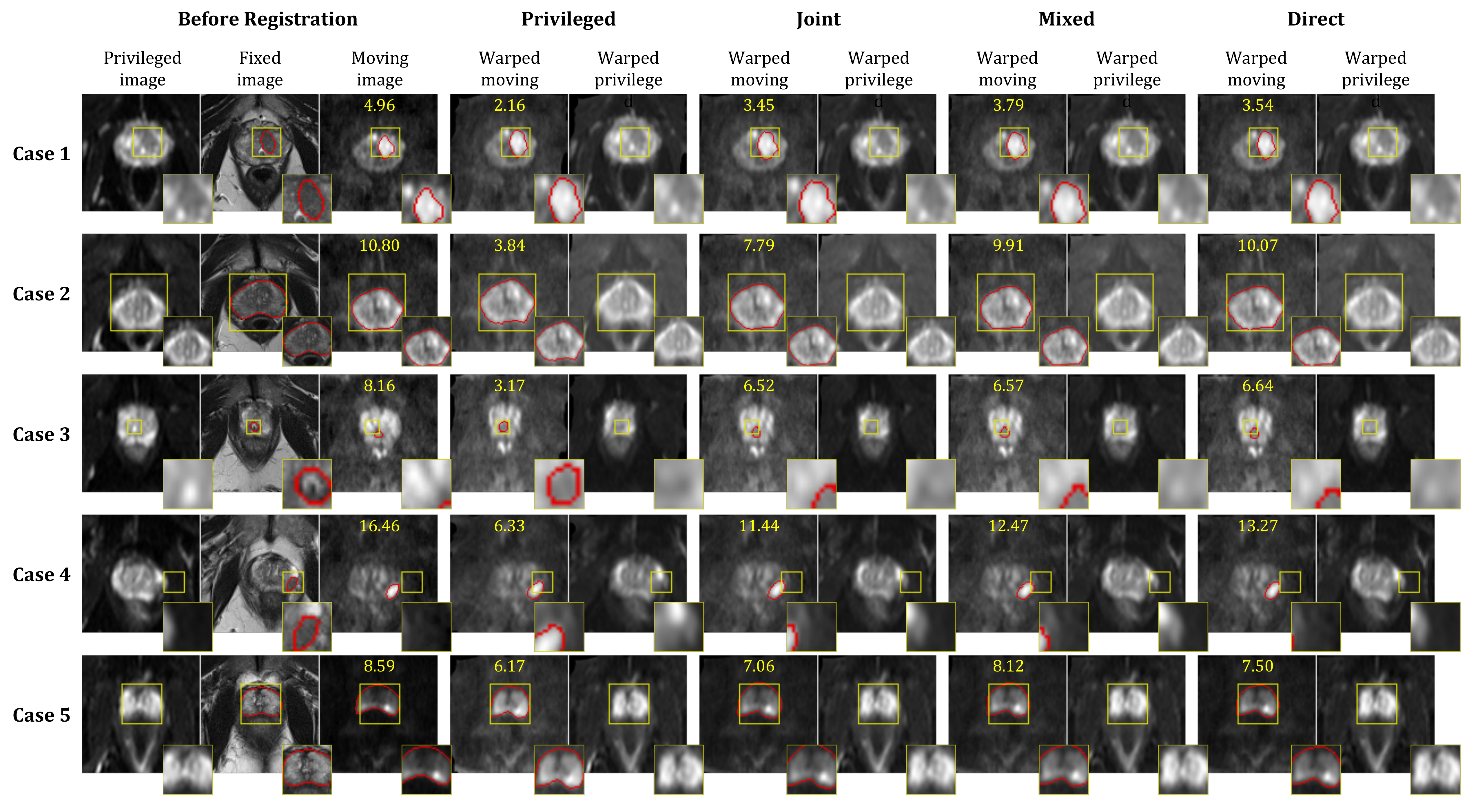}}
\caption{
Five example registered cases from different deep learning methods. The triplet in the first column indicates the privileged images, fixed images, and moving images, from left to right. The following four columns contain the results of the registered moving and privileged images, from different registration methods. The TREs(mm) from each methods are indicated in yellow, before and after registration, on the top of the moving images and the warped moving images, respectively. The yellow bounding boxes and the red validation anatomical landmarks together with their zoomed-in versions indicate ROIs for assessing registration. For each case, the bounding boxes are palced at the same reference spatial locations.}
\label{fig:comparison}
\end{figure*}

Figure~\ref{fig:comparison} provides further examples that demonstrate the potential benefits from the third modality during training, DWI$_{b=0}$ in this case. Case 1 is an example with minor misalignment. A suspected tumour was found in the central gland in the zoomed-in ROI, with the contoured tumor being aligned visually better after registration from all methods. Case 2 presents a relatively severe misalignment of the whole prostate gland, with the gland center being aligned after registration. Case 3 shows an example of a well-aligned local area of the urethra using the proposed method. Case 4 demonstrates a highly severe misalignment, for both of the prostate gland and the contoured tumor. From Case 1 to 4, it is visually recognisable that the Privileged method outperforms the others tested. For case 4, although a minor misalignment still exists after registration, the Privileged method transformed the tumor closer to the target location and shape, with respect to the reference bounding-box ROI, while the others are absent to varying degrees. Case 5 shows an example of the distortion in the gland posterior region, which was reduced by the registration.

% To further investigate the performance, the differences in TRE and MI values before and after registration are also plotted in Fig.\ref{fig:results.BAplot}. It demonstrates an effective and generalisable registration algorithm with both non-negative means from MIs and TREs, although the latter is with high variance. 

\begin{figure*}[!t]
\centerline{\includegraphics[width=2.1\columnwidth]{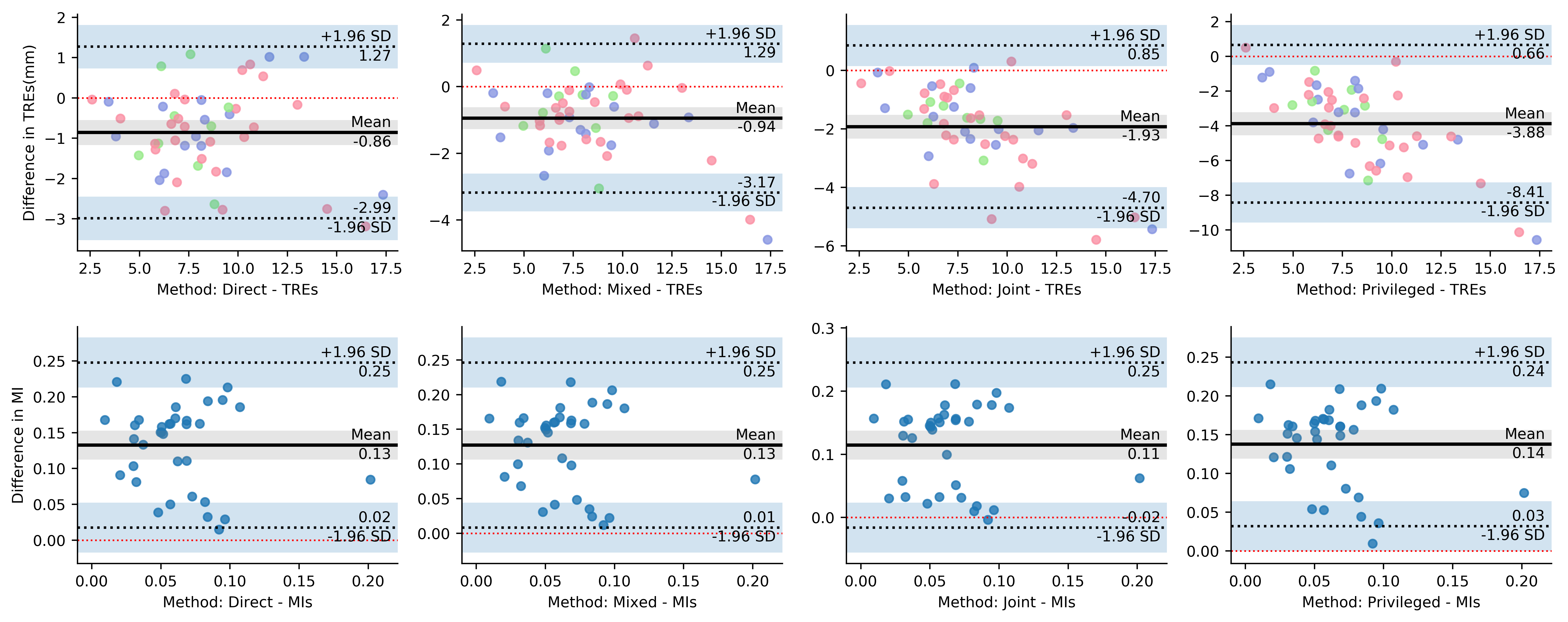}}
\caption{Bland-Altman plots of the TRE and MI differences following the proposed privileged supervision registration algorithm. Each point represents a pair of landmarks in the holdout set, where the x axis represents the TREs (1st row) and the MIs (2nd row) before registration and the y axis represents the differences of the TREs and MIs after registration. The colors of the points in the first row of figures indicates the types of the landmarks (Blue: tumors; Yellow: urethra; Red: zonal structures).}
\label{fig:results.BAplot}
\end{figure*}

To further investigate the performance, a set of Bland-Altman plots, from each proposed method, are provided to show the differences in MI and TREs before and after registration in Fig. ~\ref{fig:results.BAplot}. Each point represents a pair of landmarks in the holdout set, where the x axis represents the TRE before registration and the y axis represents the difference in TRE after registration. For each of the proposed method, improvements are observed both on MIs and TREs. The Privileged method outperforms the others with improvements of 3.88mm and 0.14, in TREs and MIs, respectively.

For the inference time, our proposed method got 0.12s while the NiftyReg got 10.19s for registering each pair of 3D image, both with GPU acceleration.

\begin{table*}[tp] 
\centering 
\caption{Holdout set performance for registering T2w-DWI$_{b=2000}$} 
\resizebox{0.8\textwidth}{!}{ 
\begin{tabular}{c|c|c|c} 
\toprule 
Methods            & Training input         & MI: Mean, Median, 90$^{th}$ Pctl. & TRE: Mean, Median, 90$^{th}$ Pctl. (mm) \\ 
\midrule 
w/o registration   &         -              & 0.063$\pm$0.034, 0.060, 0.096 & 8.331$\pm$3.016, 7.964, 11.874 \\ 
NiftyReg           & non-learning method    & 0.076$\pm$0.051, 0.068, 0.099 & 8.128$\pm$2.935, 7.862, 11.121 \\ 
Direct             & T2w-DWI$_{b=2000}$     & 0.196$\pm$0.062, 0.200, 0.289 & 7.471$\pm$3.067, 6.895, 11.969 \\ 
Mixed              & T2w-DWI$_{b=0,b=2000}$ & 0.190$\pm$0.063, 0.200, 0.281 & 7.386$\pm$2.801, 7.132, 12.110 \\ 
Joint              & T2w-DWI$_{b=0,b=2000}$ & 0.178$\pm$0.070, 0.191, 0.269 & 6.405$\pm$2.356, 6.138, 9.730 \\ 
Privileged Sup.    & T2w-DWI$_{b=0,b=2000}$ & \textbf{0.201$\pm$0.055, 0.204, 0.277} & \textbf{4.456$\pm$2.055, 4.339, 6.860} \\ 
\bottomrule 
\end{tabular}} 
\label{tab:results} 
\end{table*} 

\begin{table*} 
\centering 
\caption{Holdout set performance in 10\% and 20\% samples with the largest initial misalignment (pre-registration independent stratification) and those with the most improvement (selective results for reference). As a reference, for the registration results with the largest initial misalignment, the results before registration are MI:0.02$\pm$0.01, TREs:14.94$\pm$1.71mm (top 10\%) and MI:0.03$\pm$0.01, TREs:12.93$\pm$2.37mm (top 20\%), respectively.}
\resizebox{1.0\textwidth}{!}{ 
\begin{tabular}{c|cc|cc|cccc|cccc} 
\hline 
& \multicolumn{2}{c|}{\begin{tabular}[c]{@{}c@{}}10\% with largest\\ initial misalignment\end{tabular}} & \multicolumn{2}{c|}{\begin{tabular}[c]{@{}c@{}}20\% with largest\\ initial misalignment\end{tabular}} & \multicolumn{4}{c|}{\begin{tabular}[c]{@{}c@{}}10\% with most \\ improvement (selective)\end{tabular}} & \multicolumn{4}{c}{\begin{tabular}[c]{@{}c@{}}20\% with most \\ improvement (selective)\end{tabular}} \\ \cline{6-13} 
&                                                           &                                           &                                                           &                                           & \multicolumn{2}{c|}{Before}                                   & \multicolumn{2}{c|}{After}             & \multicolumn{2}{c|}{Before}                                  & \multicolumn{2}{c}{After}              \\ \cline{2-13}  
& \multicolumn{1}{c|}{MI}                                   & TREs(mm)                                  & \multicolumn{1}{c|}{MI}                                   & TREs(mm)                                  & \multicolumn{1}{c|}{MI}    & \multicolumn{1}{c|}{TREs(mm)}    & \multicolumn{1}{c|}{MI}   & TREs(mm)   & \multicolumn{1}{c|}{MI}    & \multicolumn{1}{c|}{TREs(mm)}   & \multicolumn{1}{c|}{MI}   & TREs(mm)   \\ \hline 
NiftyReg   & \multicolumn{1}{c|}{0.03$\pm$0.01} & 14.77$\pm$1.95 & \multicolumn{1}{c|}{0.03$\pm$0.02}  & 12.17$\pm$3.14 & \multicolumn{1}{c|}{0.07$\pm$0.03} & \multicolumn{1}{c|}{11.34$\pm$1.16} & \multicolumn{1}{c|}{0.15$\pm$0.08} & 9.42$\pm$2.02 & \multicolumn{1}{c|}{0.06$\pm$0.03} & \multicolumn{1}{c|}{10.27$\pm$3.05} & \multicolumn{1}{c|}{0.12$\pm$0.07} & 9.15$\pm$3.17  \\ 
Direct     & \multicolumn{1}{c|}{0.17$\pm$0.05} & 13.44$\pm$1.13 & \multicolumn{1}{c|}{0.16$\pm$0.04}  & 12.24$\pm$1.66 & \multicolumn{1}{c|}{0.07$\pm$0.03} & \multicolumn{1}{c|}{11.06$\pm$3.80} & \multicolumn{1}{c|}{0.28$\pm$0.03} & 8.22$\pm$3.68 & \multicolumn{1}{c|}{0.08$\pm$0.03} & \multicolumn{1}{c|}{10.12$\pm$4.14} & \multicolumn{1}{c|}{\textbf{0.28$\pm$0.03}} & 7.68$\pm$3.94  \\ 
Mixed      & \multicolumn{1}{c|}{0.16$\pm$0.05} & 12.59$\pm$0.25 & \multicolumn{1}{c|}{0.16$\pm$0.05}  & 11.66$\pm$1.20 & \multicolumn{1}{c|}{0.07$\pm$0.03} & \multicolumn{1}{c|}{12.63$\pm$4.45} & \multicolumn{1}{c|}{0.28$\pm$0.03} & 9.32$\pm$3.98 & \multicolumn{1}{c|}{0.08$\pm$0.03} & \multicolumn{1}{c|}{10.12$\pm$4.14} & \multicolumn{1}{c|}{0.27$\pm$0.02} & 7.55$\pm$3.46  \\ 
Joint      & \multicolumn{1}{c|}{0.13$\pm$0.07} & 10.99$\pm$1.15 & \multicolumn{1}{c|}{0.13$\pm$0.06}  & 9.49$\pm$1.82 & \multicolumn{1}{c|}{0.07$\pm$0.03} & \multicolumn{1}{c|}{13.63$\pm$3.20} & \multicolumn{1}{c|}{0.27$\pm$0.03} & 8.57$\pm$2.93 & \multicolumn{1}{c|}{0.08$\pm$0.03} & \multicolumn{1}{c|}{11.13$\pm$3.71} & \multicolumn{1}{c|}{0.27$\pm$0.02} & 6.99$\pm$3.08  \\ 
Privileged & \multicolumn{1}{c|}{\textbf{0.18$\pm$0.04}} & \textbf{7.45$\pm$0.88} & \multicolumn{1}{c|}{\textbf{0.17$\pm$0.03}}  & \textbf{6.77$\pm$1.36} & \multicolumn{1}{c|}{0.07$\pm$0.03} & \multicolumn{1}{c|}{13.59$\pm$3.29} & \multicolumn{1}{c|}{\textbf{0.28$\pm$0.03}} & \textbf{5.16$\pm$2.11} & \multicolumn{1}{c|}{0.08$\pm$0.03} & \multicolumn{1}{c|}{11.39$\pm$3.26} & \multicolumn{1}{c|}{0.27$\pm$0.02} & \textbf{4.07$\pm$2.09}  \\ \hline 
\end{tabular}} 
\label{tab:results.top} 
\end{table*}

\begin{table}[tp] 
\centering 
\caption{T2w-DWI$_{high-b}$ registration performance on external validation data sets.} 
\label{tab:external} 
\resizebox{\columnwidth}{!}{% 
\begin{tabular}{c|cc|cc} 
\hline 
& \multicolumn{2}{c|}{Data set A} & \multicolumn{2}{c}{Data set B} \\ \cline{2-5} 
& \multicolumn{1}{c|}{MI}  & TREs(mm) & \multicolumn{1}{c|}{MI} & TREs(mm) \\ \hline 
w/o registration & \multicolumn{1}{c|}{0.06$\pm$0.03}    & 11.01$\pm$4.06     & \multicolumn{1}{c|}{0.04$\pm$0.02}   & 6.89$\pm$2.14     \\ 
NiftyReg         & \multicolumn{1}{c|}{0.13$\pm$0.06}    & 8.67$\pm$4.71     & \multicolumn{1}{c|}{\textbf{0.16$\pm$0.05}}   & 5.34$\pm$3.42     \\ 
Direct           & \multicolumn{1}{c|}{0.24$\pm$0.08}    & 9.99$\pm$4.33     & \multicolumn{1}{c|}{0.07$\pm$0.03}   & 5.45$\pm$2.81     \\ 
Mixed            & \multicolumn{1}{c|}{0.24$\pm$0.08}    & 8.43$\pm$4.61     & \multicolumn{1}{c|}{0.07$\pm$0.03}   & 5.26$\pm$2.84    \\ 
Joint            & \multicolumn{1}{c|}{0.22$\pm$0.09}    & 9.19$\pm$4.20     & \multicolumn{1}{c|}{0.06$\pm$0.02}   & 5.98$\pm$2.42     \\ 
Privileged       & \multicolumn{1}{c|}{\textbf{0.25$\pm$0.07}}    & \textbf{6.53$\pm$4.20}     & \multicolumn{1}{c|}{0.08$\pm$0.03}   & \textbf{4.58$\pm$2.72}     \\ \hline 
\end{tabular}} 
\end{table}

\subsection{The need for registering T2w and \texorpdfstring{DWI$_{b=2000}$}{DWIb=2000}}
The MI and TREs on the test data set are computed to indicate the original difference between the two images without registration. All registration methods have made positive contributions to align images based on the increased MI values. All of the tested registration methods reduced TREs. This is an indication that registration in general would help align the T2w and DWI$_{b=2000}$ scans in this application. 

Table~\ref{tab:results.top} provide results from the 10\% and 20\% cases with the largest initial misalignment and 10\% and 20\% cases with the most improvement observed after registration. The results from both these subgroups showed a larger initial misalignment and arguably more substantial improvements from the registration. For example, for 20\% cases with largest initial misalignment, the proposed privileged supervision network improved the mean TREs from 12.93 mm to 6.77 mm.

We also report a set of selective results \textit{only} for inspecting the extent of the registration error, from the 10\% and 20\% cases with the most improvement by registration, measured by TRE. However, identifying either of these scans that need registration the most remains an interesting open research question, as it may be that, based on the clinical data set used in this work, the proposed method would be of increased clinical value when applied to this subset of patient studies.

\subsection{Comparison to other privileged learning methods}
From both Table~\ref{tab:results} and Table~\ref{tab:results.top}. The proposed methods outperformed the alternative joint training and mixed sampling methods, in terms of TREs. The advantage is both consistent and statistically significant (p-values$<$0.001). This set of results demonstrate the effectiveness of the proposed privileged learning method to align the T2w and DWI$_{b=2000}$, in this application. These results concludes that adding images from a different modality to training may not be trivial and, without appropriate adaptation, may reduce the registration performance. In addition, the same network was trained with the Privileged method with 10 random affine transformations in the Monte-Carlo resampling (i.e., $I=10$), described in Sec.~\ref{sec:methods.surrogate}. The mean TRE increased from 4.456$\pm$2.055mm to 4.347$\pm$1.845mm. Improvement was observed but without statistical significance (p-value=0.34). Considering the computation-associated feasibility of the propose algorithm, we chose to report results based on $I=5$, as the number for the Monte-Carlo resampling.

\subsection{Comparison to other learning-based registration}
Direct registration marginally lowered the mean TRE, although with significance (p-value$<$0.001). The proposed privileged learning method obtained a lower mean TRE, compared to the direct registration method with statistical significance (p-value$<$0.001). It is consistent with the observations from the qualitative results in ~\ref{fig:comparison}, which indicates the privileged learning method showed effective registration itself (Sect~\ref{sec:results.holdout}) whilst the Direct method did not. Interestingly, the Direct method produced a relatively high MI. This may be expected as the direct algorithm was trained to maximise MI directly, but the optimization was influenced by the heavy noise can lead to inferior TREs without the potential benefits from the added DWI$_{b=0}$ images, as discussed in Sect~\ref{sec:results.holdout}. It is also interesting to report that, using T2w and DWI$_{b=0}$ as network input both in training and testing (Section~\ref{sec:exps.learning}), the warped DWI$_{b=2000}$ also led to a mean TREs of 4.59$\pm$2.01mm, outperforms the Direct, Joint, and the Mixed methods (p-value$<$0.001). These results summarise that the non-trivial difficulties in direct registering T2w and DWI$_{b=2000}$.

\subsection{Comparison to non-learning registration}
NiftyReg also improved the mean with statistical significance achieved(p-values=0.03), but the improvement is very limited. The mean and median TRE from privileged supervision is improved over those from NiftyReg results (p-value$<$0.001). Results from the two subgroups are summarised in Table~\ref{tab:results.top}. It may be interesting to report that, the selective group (the lower two columns in Table~\ref{tab:results.top}), on which registration provided most improvement, has a larger misalignment with the privileged modality, compared to those with NiftyReg. This perhaps indicates the potential utilisation of the extra anatomical and pathological information retained in the privileged images. 

\subsection{Interpretation of the external validations}

Table ~\ref{tab:external} summarises the registration performance from each method on two external validation data sets. On both data sets, our proposed Privileged method outperforms the results before registration and from the other methods (all p-values$\le$0.01). Compared with the original data set, the Data Set A is with larger initial misalignment, with a mean TRE of 11.01mm. Although Data Set B is with smaller initial misalignment, it was acquired with larger difference in acquisition protocols. For example, the MRIs from Data Set B were taken with endorectal coil and the b-value of the DWI$_{high-b}$ is only 1400 $sec/mm^2$. The Direct method and the Joint method show smaller improvements on TREs for both data sets, compared with the proposed method, albeist arguably smaller difference in the optimised MI. The Mixed method achieved relatively competitive performance on both data sets, second to the Privileged method. It probably because that the mixed sampling introduced more training data and thus increased its generalisability. It is also interesting to report that, the NiftyReg achieved lower TREs than the Direct and Joint methods in the external validation, although statistical significance was not found in these cases (p-values=0.83 and 0.27, respectively).

\section{Discussion}
The proposed use of the third modality was not only evident in helping many cases in our application, but also provide an interesting new mechanism beyond improving registration performance, by bringing in a potentially more intuitive and radiologically-interpretable modality, for future registration studies, such as investigating registration error distribution, local loss function design and evaluation methodologies.

This work examined a particularly challenging cross-modality task in registering T2w and high b-value diffusion scans, from prostate cancer patients. During the investigation, we summarise the difficulties as follows: 1) high variance exist in both imaging and validation landmark annotating, in particular, clinical data contain variable and unknown misalignment from different patients; 2) the lack of consistent and robust similarity measures as a loss function between the two complementary imaging modalities.

Largely motivated by the high efficiency from the recent learning-based registration methods, we developed and compared registration networks and their associated training strategies. More interestingly, we proposed to use a third modality image that is arguably "closer" to both images to register to help the training procedure. In experimental results, we show that such addition could indeed help the registration in a number of scenarios, with consistent and statistically significant advantages with the moderately-sized multimodal image data set from clinical practice.

In summary, the presented experimental results confirmed that the proposed registration network training method can benefit from an additional modality during training. The improvement over other learning-based method, with different ways to make use of the ``privileged modality'' or without using it at all, is effective and consistent, especially for a subset of these patient cases that with largest misalignment, therefore needing the registration the most. 

We have demonstrated the proposed registration method using a privileged modality with the specific prostate cancer imaging application. While this method has potentials for training registration networks using other types of available images in wider clinical applications, including and beyond those potential applications discussed in Section I, these require further investigation and validation.

\section{Conclusion}
We have proposed strategies for the third modality images to aid the training of bi-modality image registration networks. The competitive registration accuracy has been experimentally demonstrated on mpMR data from prostate cancer patients. The proposed novel methodology may be generally applicable to a wide range of clinical image registration tasks.

\section*{Acknowledgment}
This work was supported by the International Alliance for Cancer Early Detection, a partnership between Cancer Research UK [C28070/A30912; C73666/A31378], Canary Center at Stanford University, the University of Cambridge, OHSU Knight Cancer Institute, University College London and the University of Manchester. This work was also supported by the Wellcome/EPSRC Centre for Interventional and Surgical Sciences [203145Z/16/Z], the Wellcome/EPSRC Centre for Medical Engineering 203148/Z/16/Z; NS/A000049/1] (TV), the EPSRC CDT in i4health [EP/S021930/1], an MRC Clinical Research Training Fellowship [MR/S005897/1] (VS), a Royal Academy of Engineering / Medtronic Research Chair [RCSRF1819\textbackslash7\textbackslash734] (TV). For the purpose of Open Access, the author has applied a CC BY public copyright licence to any Author Accepted Manuscript version arising from this submission.

\bibliographystyle{IEEEtran}
\bibliography{tmi.bbl}
\end{document}